\documentclass{article}

     \usepackage[final]{neurips_2021}

\usepackage[utf8]{inputenc} 
\usepackage[T1]{fontenc}    
\usepackage{hyperref}       
\usepackage{url}            
\usepackage{booktabs}       
\usepackage{amsfonts}       
\usepackage{nicefrac}       
\usepackage{microtype}      
\usepackage{xcolor}         
\usepackage{graphicx}
\usepackage{makecell}

\title{Rethinking Generalization Performance of Surgical Phase Recognition with Expert-Generated Annotations}

\author{%
  Seungbum Hong, Jiwon Lee, Bokyung Park, and Min-Kook Choi\thanks{Corresponding author.} \\
  hutom\\
  Seoul, Republic of Korea\\
  \texttt{\{qbration21, jiwon, bokyung, mkchoi\}@hutom.io} \\
  \AND
  Woo Jin Hyung \\
  hutom, Yonsei University College of Medicine \\
  Seoul, Republic of Korea \\
  \texttt{wjhyung@hutom.io} \\
  \And
  Ahmed A. Alwusaibie, Anwar H. Alfadhel, and SungHyun Park  \\
  Yonsei University College of Medicine \\
  Seoul, Republic of Korea \\
  \texttt{GODON@yush.ac} \\
}

\begin{document}

\maketitle

\begin{abstract}
As the area of application of deep neural networks expands to areas requiring expertise, e.g., in medicine and law, more exquisite annotation processes for expert knowledge training are required. In particular, it is difficult to guarantee generalization performance in the clinical field in the case of expert knowledge training where opinions may differ even among experts on annotations. To raise the issue of the annotation generation process for expertise training of CNNs, we verified the annotations for surgical phase recognition of laparoscopic cholecystectomy and subtotal gastrectomy for gastric cancer. We produce calibrated annotations for the seven phases of cholecystectomy by analyzing the discrepancies of previously annotated labels and by discussing the criteria of surgical phases. For gastrectomy for gastric cancer has more complex twenty-one surgical phases, we generate consensus annotation by the revision process with five specialists. By training the CNN-based surgical phase recognition networks with revised annotations, we achieved improved generalization performance over models trained with original annotation under the same cross-validation settings. We showed that the expertise data annotation pipeline for deep neural networks should be more rigorous based on the type of problem to apply clinical field.
\end{abstract}

\section{Introduction}

Because algorithms using pattern recognition or machine learning are applied for various tasks, problem definition and data acquisition are as important as the model's performance [1-4]. Training using deep neural networks, a trend in machine learning, requires large amounts of annotated data to avoid overfitting with a large number of parameters. Crowdsourcing-based tools, such as Amazon Mechanical Turk, which requires simple training for annotators, can be widely used to efficiently collect large amounts of data, especially for visual tasks with high data redundancy and diverse variations between images [5-9]. The process of securing the annotation database in the visual recognition problem is essential for defining and training algorithms but is considered separately from the model training process.

\begin{figure*}[t!]
\includegraphics[width=\textwidth]{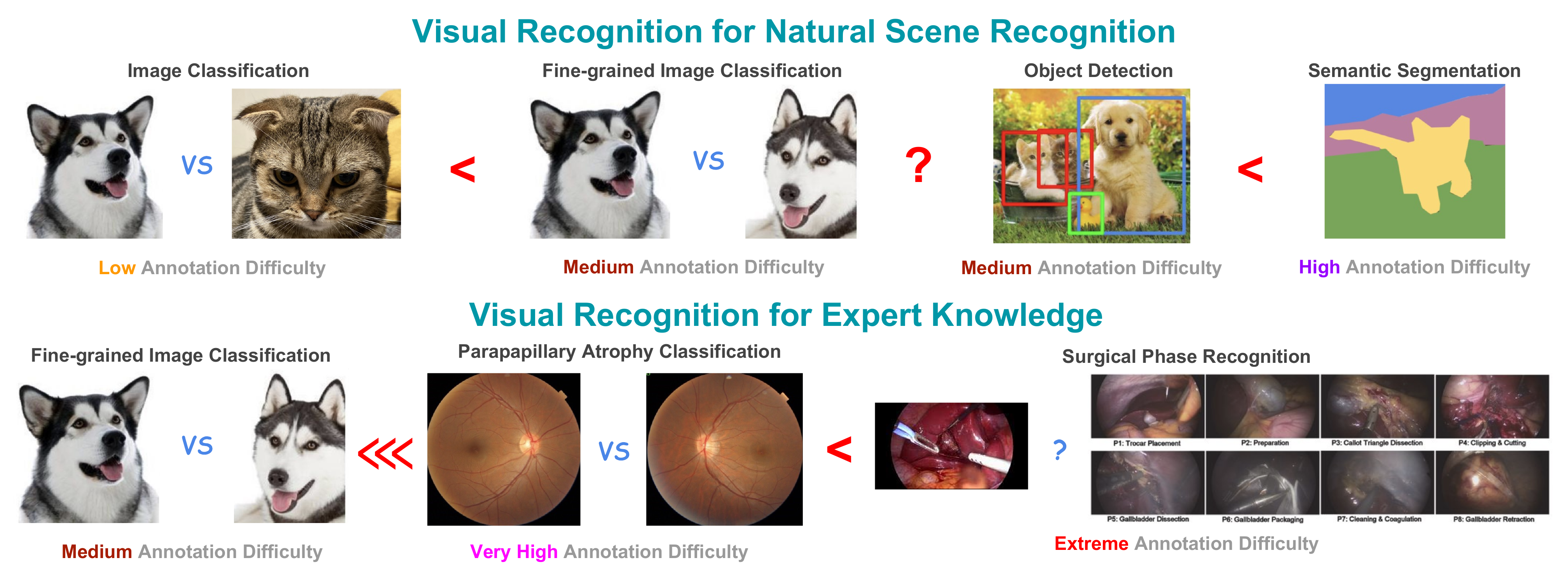}
\caption{\textbf{Annotation difficulty of expert knowledge}. Depending on the annotation difficulty, thorough verification of the data collection and annotation process is required. We exemplify the surgical phase recognition problem for laparoscopic cholecystectomy and argue that a more difficult problem requires more consideration in the annotation process.} 
\label{fig1}
\end{figure*}

Figure \ref{fig1} shows an example of the relationship between the annotation difficulty and the visual recognition problem. In the example in Figure \ref{fig1}, if the effort to annotate a single sample, e.g., image classification or object detection, is relatively small and high generalization performance can be easily achieved, the difficulty level of the annotation is relatively low. However, even in the image classification problem, assuming that the classifier classifies Malamute and Siberian Husky if the annotator lacks prior knowledge of the dog's breed, the difficulty of annotating may increase, increasing the overall difficulty level of the generalization. It is important to analyze the difficulty level of the annotation because the range of visual recognition problems to be solved using deep learning is expanding from simple recognition to problems requiring expertise. In particular, problems involving medical diagnosis, aid, or surgery are often impossible to annotate based on appearance only [10-17], e.g., through visual recognition of natural scenes. Furthermore, the annotation of supervision for visual recognition that includes this expertise is very expensive, so the inspection effort is also very expensive. 

In this paper, we argue that when deep neural networks train expert knowledge such as surgical phase recognition which has a very high annotation difficulty, one needs to consider the more precise annotation procedure than the typical visual recognition problem. Our analysis shows that surgical phase recognition can have very different generalization performance depending on the type of supervision, which means that it can have a very large bias when applying the training model in the clinical field. To do this, we compared the generalization performances of the same model trained with revised annotations obtained from different experts' consensus processes. we tried two different datasets generation pipeline for surgical phase recognition [17-23] of laparoscopic cholecystectomy and robotic subtotal gastrectomy for gastric cancer. To reconsider the accuracy of the existing annotations public dataset for surgical phase recognition of laparoscopic cholecystectomy, we used 24 surgery videos from the Cholec80 [17] dataset and we matched the criteria of two specialists and created two additional revised annotations. At the same time, we devised a pipeline capable of generating consensus annotations for 40 cases of surgery by the Da Vinci system to agree with experts in gastrectomy for gastric cancer, one of the most complex surgeries. To best our knowledge, This is the first approach to automated surgical phase recognition of subtotal gastrectomy for gastric cancer. By comparison of validation performance with surgical phase recognition models based on Convolutional Neural Networks (CNN), we confirmed the nontrivial differences in the test results of the models trained under the same cross-validation conditions except the annotation. 

\section{Background}
\smallskip\noindent \textbf{Medical data annotation for visual recognition.} Annotations for training visual recognition models in natural scenes are primarily made to achieve classification, detection, and segmentation of images and videos. Annotations are sometimes performed to train algorithms for regression problems, such as pose recognition and image registration. They are generally annotated at a level that allows training by the public with a mild iterative training process. For this reason, crowdsourcing platforms [5-9] that use web-based annotation tools are widely used in visual recognition problems. 
As the performance of visual recognition technology improves, various annotation tools for training are being introduced in the field of medical imaging  [10, 11]. In many cases, X-ray, magnetic resonance imaging, and computed tomography images are used as annotation targets for automated diagnosis or diagnostic assistance [12-15]. As the demand for data for training increases, web-based annotation tools or methodologies, such as crowdsourcing, for medical imaging are also being introduced [16]. Recently, automated visual recognition has started to be used as a medical analysis method for surgical videos, and recent datasets such as Cholec80 [17] have been widely used. Annotation for surgical phase recognition with differences from other visual recognition problems is very expensive because it requires expertise in a specific type of surgery. Annotations made by different experts are likely to be different because frame-level annotations must be performed for all surgical procedures, and at the same time, surgical phases of a particular duration must be determined by considering the continuous information on the temporal dimension. In this study, we show that the generalization performance for automated surgical phase recognition varies considerably with expert annotation. This argues that more attention should be paid to the annotation process considering its use in the clinical field.

 \smallskip\noindent \textbf{Surgical phase recognition.} With the development of visual recognition technologies, various surgical phase recognition algorithms have been developed to analyze the surgical process using surgical videos. Hierarchical hidden Markov model with two levels to process temporal information of features encoded by CNNs is developed by [17]. The features obtained from the hierarchical Hidden Markov model were used to achieve high accuracy surgical phase recognition for laparoscopic cholecystectomy using a support vector machine classifier. In addition, an efficient training method from a small number of surgery videos using a CNN-LSTM [27] structure with the same dataset used in [17] has been proposed [18]. In [19], CNN-LSTM-based surgical phase recognition was performed with the Cholec120 dataset, which was extended from the Cholec80 dataset, using the pre-trained remaining surgery duration network for self-supervised learning. In addition to cholecystectomy using the Cholec80 dataset [20], surgical phase recognition for laparoscopic sigmoidectomy [21] and laparoscopic hysterectomy [22] was performed. In [23], CNN-based surgical phase recognition was performed using prostatectomy videos for robotic surgery. 

 \smallskip\noindent \textbf{Saptiotemporal encoding for visual recognition.} Various efforts have been made to efficiently encode spatiotemporal feature in the visual recognition problem [24, 25]. In the early days of the deep learning era, models were proposed to utilize 2D-CNN feature and motion information together [26], and a 2D-CNN-LSTM model including the consideration of the temporal domain of 2D-CNN features was proposed [27]. Recent researches have been reported that a model that simultaneously encodes spatiotemporal information in one network may perform better than explicitly encoding separately processed motion information [28-32]. Our goal is not to suggest a good spatiotemporal model, but to see how the state-of-the-art performance models influence the results of inference in training expertise under different annotations. For this purpose, the results of surgical phase recognition were verified with models based on 2D-CNN-LSTM [27] and 3D-CNN architectures [29, 31].

\section{Annotation verification}
\begin{figure*}[t!]
\begin{center}
\includegraphics[width=0.9\textwidth]{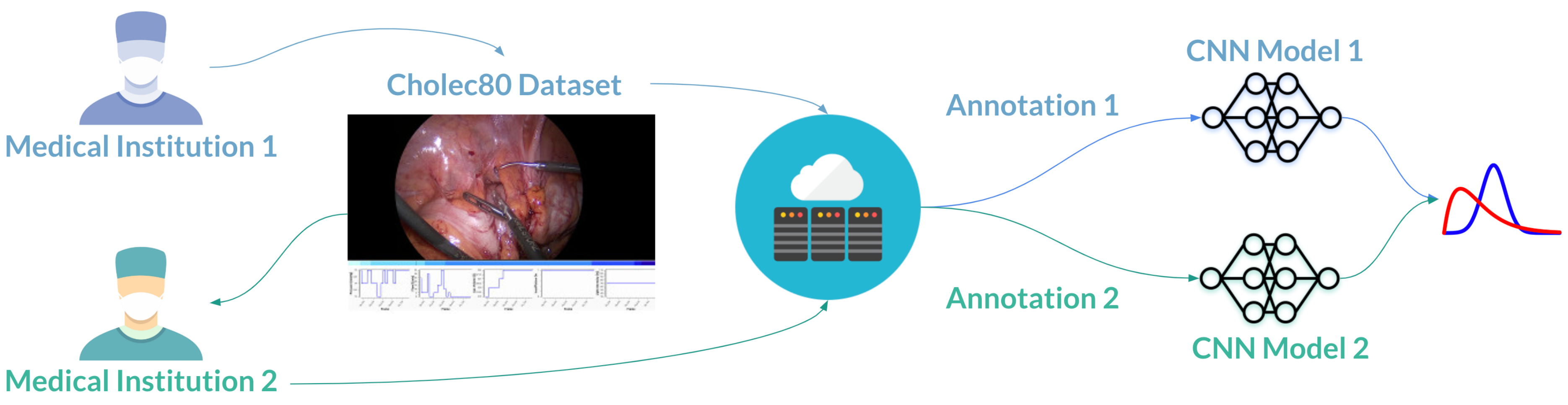}
\caption{\textbf{Schematic of annotation verification process for surgical phase recognition in laparoscopic cholecystectomy.} With 24 surgical videos of cholecystectomy, multiple expert opinions were processed to validate ground truth annotations. We trained the CNN-based classification models with the same architecture and initial weights as the original and revised annotations and verified whether there is a change in inference performance.} 
\label{fig2}
\end{center}
\end{figure*}

\begin{figure}[t!]
\begin{center}
\includegraphics[width=0.40\linewidth]{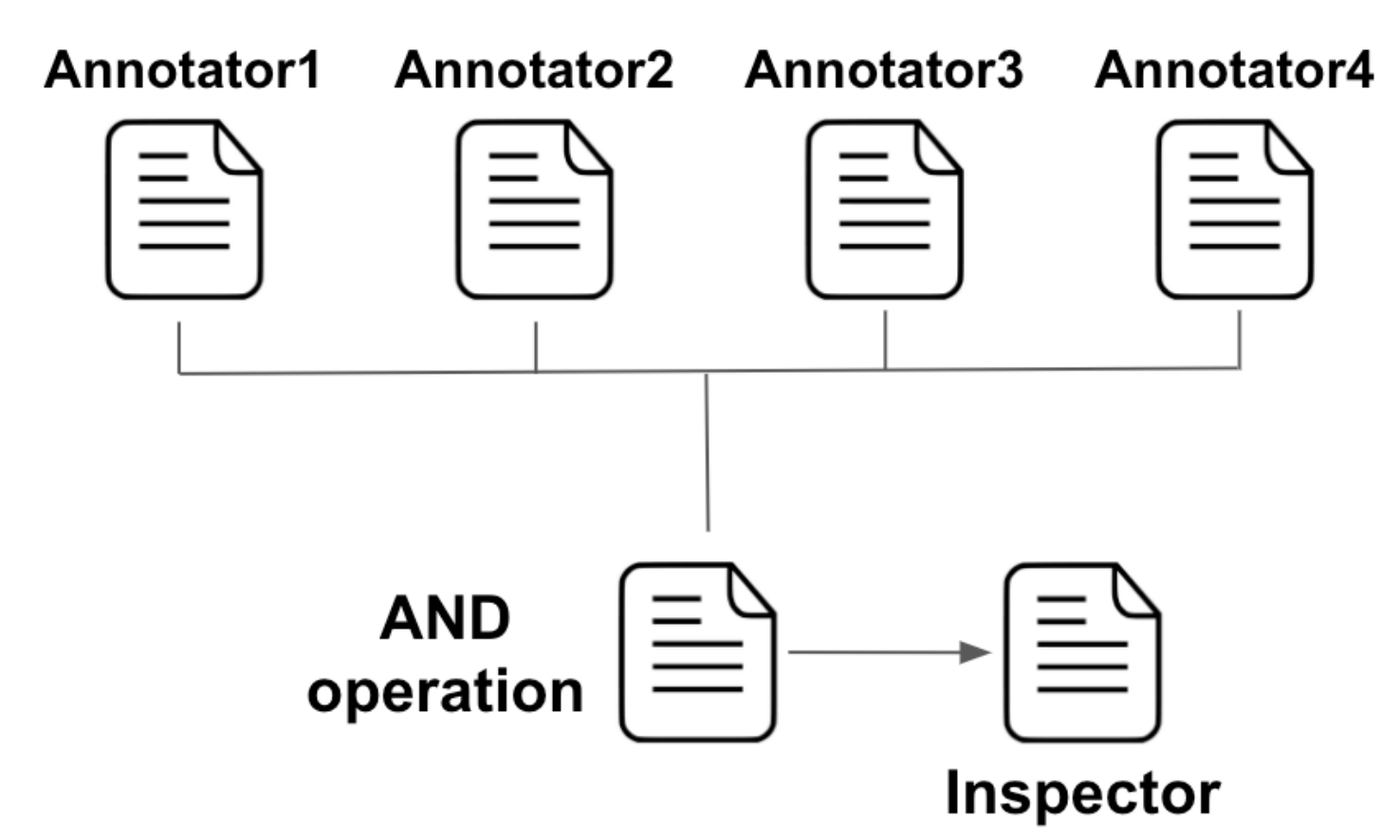}
\caption{\textbf{Schematic of generating consensus annotation for surgical phase recognition in robotic subtotal gastrectomy for gastric cancer.} Since gastrectomy for gastric cancer is one of the most complicated surgeries, we have established an additional process for annotation generation. A total of 4 specialists respectively annotate for each surgical video and generate a new label through an AND operation between the annotations. For frames that remain blank after the AND operation, another specialist writes the final label by colligating the opinions of other annotations as an inspector.}
\label{fig3}
\end{center}
\end{figure}

We used a surgical phase recognition dataset of laparoscopic cholecystectomy to perform the annotation verification process. The original videos of the laparoscopic cholecystectomy were a subset of the Cholec80 dataset [17]. We selected 24 videos released for the training surgical phase recognition. The revised annotations of the 24 videos were generated through expert consensus with two specialists. In addition to laparoscopic cholecystectomy, we built an annotation generation pipeline for gastrectomy for gastric cancer in 40 cases to try to recognize more complex surgery\footnote{Please refer to supplementary materials for more details of the refined surgical phase and surgical phase for subtotal gastrectomy for gastric cancer.}. The annotation and verification process went through the processes shown in Figures \ref{fig2} and \ref{fig3}, and a total of 5 specialists participated in this process.

\noindent\textbf{Surgical phase recognition in laparoscopic cholecystectomy.} Seven phases of laparoscopic cholecystectomy surgery were defined in the Cholec80 dataset [17]. The preparation, Calot triangle dissection, clipping and cutting, gallbladder dissection, gallbladder packaging, cleaning and coagulation, and gallbladder retraction were indicated with index 0, 1, 2, 3, 4, 5, and 6, respectively. The annotation criteria were designed to target the multiclass classification problem. Labels were annotated in a form in which all frames were extracted from a video recording with a specific case of surgery. Depending on the progress of the surgery, the surgical phase may have been preceded by a higher index phase or returned to the previous phase, but multiple labels for one frame were not allowed. Depending on the progress of the surgery, the entire surgery length was distributed within approximately 30 min to 2 h. Currently, only one label is provided as the correct answer for the classification dataset. We used the provided annotations from Cholec80 as Annotation 1 to train the CNN-based surgical phase recognition model.

\begin{figure*}[t!]
\begin{center}
\includegraphics[width=0.85\textwidth]{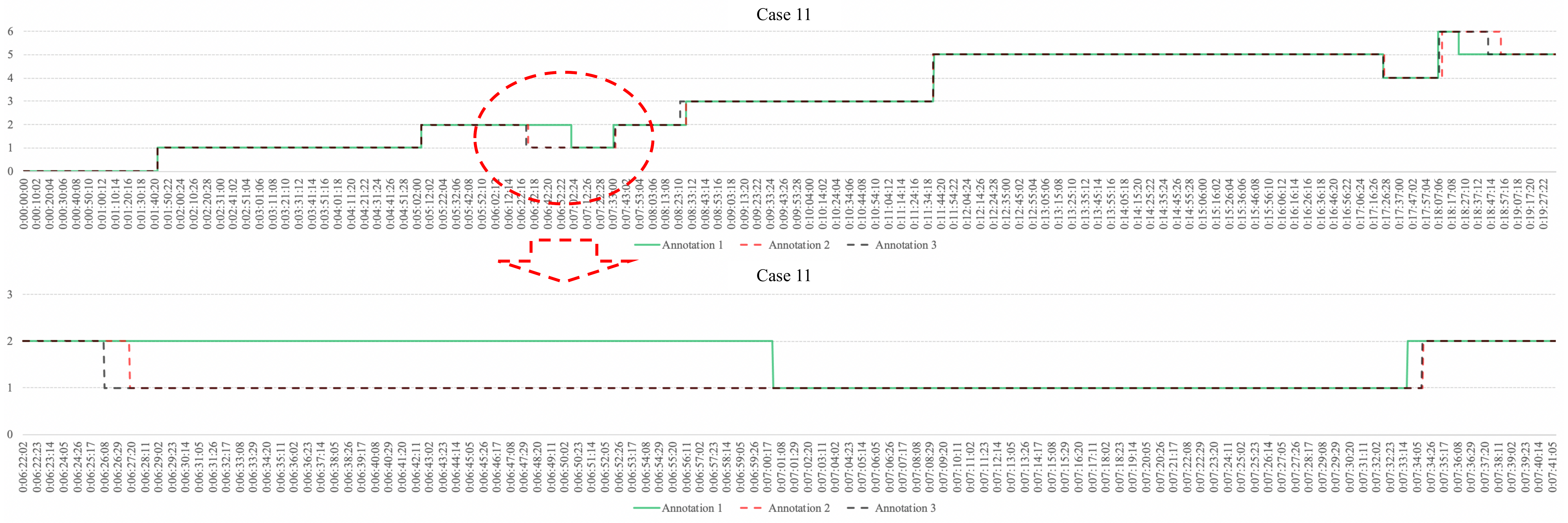}
\caption{\textbf{Comparison of annotations before and after expert consensus.} (a) Plot of the existing and revised annotations on the original video at the same timestamp. (b) Result of re-plotting only the section where the change occurs in a specific surgical phase in the video. Disagreements can be seen near the surgical phase changes.} 
\label{fig4}
\end{center}
\end{figure*}

\noindent\textbf{Surgical phase recognition in robotic subtotal gastrectomy.} We divided the gastrectomy surgical phases for gastric cancer into 27 phases, based on ARMES [32]$^1$. ARMES is a system that divides surgical processes into 20 separate phases, from start to finish, to provide the surgeon with information on gastrectomy for gastric cancer. We added one phase to the 21-phases surgical procedure, as defined by ARMES, and composed 27 surgical steps. This included the 21 surgical procedures as part of the actual gastrectomy and six non-gastrectomy procedures. At this time, the six non-gastrectomy procedure types include actions that are irrelevant to gastrectomy, such as camera cleaning, adhesiolysis, and facility management, which may appear during gastrectomy. Since such actions are unrelated to the gastrectomy and do not depend on a specific surgical step; they may affect the training process of the surgical phase recognition model. Therefore, these were included separately.

\noindent\textbf{Annotation revision for laparoscopic cholecystectomy.} We revised the annotation of the phase of the laparoscopic cholecystectomy surgery through a consensus process of three specialists. Figure \ref{fig4} shows an example of the agreed annotation of each surgical phase, which was updated based on the original annotation. The consensus among specialists for the seven steps of cholecystectomy was based on their consensus, and a separate consensus process was performed on the points where changes in the surgical phase occurred or where exceptions occurred. Finally, two types of annotation (Annotation 2 and 3) revised by two specialists were created from Annotation 1. Figure \ref{fig4} (a) shows the difference between the revised annotation and the original annotation in the entire cholecystectomy, and (b) shows the difference between the annotations according to the opinions of the experts in the section where the surgical phase changed.

\noindent\textbf{Annotation pipeline for robotic subtotal gastrectomy.} For accurate labeling for more complex surgery, we performed frame-level annotation for 40 cases that were cross-validated by five specialists. For cross-validation, we received annotated information according to each opinion from four specialists, and this information was compared with the areas where the opinions matched by AND operation. For labels containing blank generated by the AND operation, another specialist annotates the label as an investigator, resulting in the final consensus label. Robotic surgery for 40 cases of gastric cancer proceeded subtotal gastrectomy and nineteen cases were recorded on the da Vinci Si system, while twenty-one videos were recorded on the da Vinci Xi. Figure \ref{fig3} shows a schematic illustration of the annotation verification process.

\section{Experimental Results}
To verify the difference in the model performance as the annotation changes, we trained the 2D-CNN-LSTM model [27] and the two types of 3D-CNN based models using the original and revised annotations for the Cholec80 and robotic gastrectomy for gastric cancer\footnote{Additional results and analysis are included in supplementary materials.}. As the 3D-CNN model, we selected and trained the latest-performance 3D-ResNet [28] and ECO [31] models. Test scenarios were constructed according to the structural characteristics of the trained model, and no pre- or post-processing of the input data and inference output was performed to accurately analyze the effect of the label on the model training. However, in the training and inference of each CNN model, basic data augmentation was performed as part of the regularization at the input stage of the architecture. All models have been implemented using the PyTorch v1.2.0 [33] and trained on NVIDIA V100.

The loss function for the final cross-entropy output in the recognition of all surgical stages is

\begin{equation}
  L_{CE}=-\frac{1}{M}\sum^{M}_{i}\sum^{C}_{j}t_i \log(h(x^i_j;\theta)),
\end{equation}

\noindent where $M$ is the size of the input minibatch, and  $C$ is the total number of surgical phases of a each surgery ($C=7$ for cholecystectomy, 
$C=27$ for subtotal gastrectomy); $h(x^i_j;\theta)$ is the softmax output for the input sequence (or image) $x$ in each architecture with $\theta$ as the training weight, and $t$ as the ground truth label for each $x^i_j$.

\noindent\textbf{Training 2D-CNN-LSTM.} ResNet34 [34] pre-trained with ImageNet was used as the 2D-CNN model. The input clip sequence for training received 16 frames at a sampling rate of 1f/s and consisted of a total of 16 s of frame data; each input clip sequence had an overlap ratio of 0.5 with the previous input clip. ADAM was used as the optimizer; the initial learning rate was $1 \times10^{-4}$, and a scale of 0.1 was applied to the learning rate every 30 epochs. A total of 100 epochs were trained with 256 mini-batches, and the input frame was resized to 256$\times$256. For data augmentation during training, spatial random crop and left and right flips for multiple scales were applied.

\noindent \textbf{Training 3D-ResNet.} 3D-ResNet extends the residual learning structure [29] designed for efficient training of CNNs into three dimensions by considering time-level information simultaneously. We utilized a 34-layer 3D-ResNet with pre-trained by the Kinetics dataset [35] to efficiently train spatiotemporal information about the surgical phase in the surgery video. The input clip was composed of 16 frames at 1f/s as in the case of the 2D-CNN-LSTM. SGD was used as the optimizer with initial learning of 0.001, the momentum of 0.9, and weight decay of $5 \times 10^{-5}$; the learning scheduler was applied to reduce the learning rate at the plateau. The mini-batch trained a total of 200 epochs at 128; the spatial size of the input sequence was resized to 256$\times$256 and finally processed to a training input through random cropping to a size of 224$\times$224.

\noindent \textbf{Training ECO.} ECO has two structurally arranged networks to simultaneously utilize the advantages of encoding the spatial information of the 2D CNNs and the temporal information of the 3D CNNs. The 3D CNNs with a relatively large number of parameters were applied in the final output stage of the network to maximize the accuracy and efficiency of the inference speed [31]. We used pre-trained BN-Inception [36] from the ImageNet [5] as the 2D CNN for training ECO, and an 18-layer 3D ResNet network pre-trained with the Kinetics dataset [35] as the 3D CNNs. SGD with Nesterov momentum was used as the optimizer, and dropouts were applied to each fully connected layer. The initial learning rate was set to 0.001, and the learning scheduler applied a scale of 0.1 when verification errors are saturated every four epochs. We set the momentum to 0.9, weight decay to $5\times10^{-4}$, and the size of the mini-batch to 32. The input frame was resized to 224$\times$224 after applying a fixed corner crop and horizontal flip to resized frame to 240$\times$320.

\begin{table*}[t!]
  \caption{AP for the inference of surgical phase recognition for a cross-validation set of laparoscopic cholecystectomy for each annotation and model.}
  \label{tab1}
  \resizebox{\linewidth}{!}{
  \begin{tabular}{c|c|c|c|c|c|c|c}
\hline
 - & \makecell{Split 1\\ (Ann 1/2/3)}  & \makecell{Split 2\\ (Ann 1/2/3)}   & \makecell{Split 3\\ (Ann 1/2/3)}  & \makecell{Split 4\\ (An 1/2/3)}  & \makecell{Split 5\\ (Ann 1/2/3)}  & \makecell{Split 6\\ (Ann 1/2/3)}  & \makecell{mAP \\ (Ann 1/2/3)} \\ \hline
2D-CNN-LSTM & 65.8/66.1/62.3  & 59.2/60.4/58.5 & 51.1/37.8/51.59 & 51.7/48.1/57.1 & 70.9/70.3/68.0 & 67.8/\textbf{72.0} /68.6& 61.1/59.1/61.0 \\ \hline 
3D-ResNet & 67.8/73.2/72.9  & 49.9/61.3/60.6 & 52.3/51.1/\textbf{57.1} & 52.5/53.1/46.3 & \textbf{73.5}/68.7/72.9 & 66.7/65.1/62.9 & 60.5/61.5/\textbf{67.6}  \\ \hline 
ECO & 74.1/\textbf{75.0}/69.1 & 64.4/\textbf{70.9}/65.5 & 45.6/49.1/46.9 & 57.3/\textbf{60.3}/55.9 & 71.5/71.5/\textbf{73.5} & 69.6/67.7/68.6 & 63.8/65.7/63.2  \\ \hline
\end{tabular}
}
\end{table*}

\begin{table*}[t!]
  \caption{Averaged AP for inference of the surgical phase recognition of robotic gastrectomy for gastric cancer with each annotation and model.}
  \label{tab2}
  \begin{center}
\scalebox{0.8}{
  \begin{tabular}{c|c|c|c|c|c|c|c|c|c}
\hline
2D-CNN-LSTM & Ann1  & Con-Ann1 & Ann2 & Con-Ann2 & Ann3 & Con-Ann.3 & Ann4 & Con-Ann4 & Con \\ \hline 
Split1 		& 65.07  & +2.49 	& 63.23 & +4.33 	& 67.38 & +0.18 	& 64.77 & +2.79 	& \textbf{67.56}  \\ \hline 
Split2 		& 59.32  & +4.81 	& 62.21 & +1.92 	& 60.21 & +3.92 	& 55.72 & +8.41 	& \textbf{64.13}  \\ \hline 
Split3 		& 68.59  & +3.05 	& 68.8 & +2.84 		& 67.59 & +4.05 	& 68.57 & +3.07 	& \textbf{71.64}  \\ \hline  \hline
3D-ResNet 	& Ann1  & Con-Ann1 & Ann2 & Con-Ann2 & Ann3 & Con-Ann.3 & Ann4 & Con-Ann4 & Con \\ \hline 
Split1 		& 68.31  & +1.98 	& 67.52 & +2.77 	& 67.54 & +2.75 	& 65.34 & +4.95 	& \textbf{70.29}  \\ \hline 
Split2 		& 66.84  & +3.11 	& 64.68 & +5.27 	& 68.29 & +1.66 	& 62.46 & +7.49 	& \textbf{69.95}  \\ \hline 
Split3 		& 72.4  &  +4.79	& 72.25 & +4.94 	& 74.31 & +2.88 	& 74.16 & +3.03 	& \textbf{77.19}  \\ \hline  \hline
ECO 		& Ann1  & Con-Ann1 & Ann2 & Con-Ann2 & Ann3 & Con-Ann.3 & Ann4 & Con-Ann4 & Con \\ \hline 
Split1 		& 64.97  & +4.69 	& 63.89 & +5.77 	& 66.39 & +3.27 	& 67.13 & +2.53 	& \textbf{69.66}  \\ \hline 
Split2 		& 60.46  & +4.76 	& 61.85 & +3.37 	& 61.26 & +3.96 	& 60.5 &  +4.72 	& \textbf{65.22}  \\ \hline 
Split3 		& 71.45  & +4.1	 	& 70.73 & +4.82 	& 72.79 & +2.76 	& 72.28 &+ 3.27 	& \textbf{75.55}  \\ \hline  \hline
\end{tabular}
}
\end{center}
\end{table*}

\noindent \textbf{Testing.} The inference scenario for obtaining the testing performance received a sequence of the same size as that of the training process as input and output the final confidence value for each surgical phase. In the 2D CNN-LSTM and ECO, the spatial feature information encoded in the 2D-CNN was used as input from the previous  $(k-15)$ frames to the $k$th frame for inference on the $k$th frame; in the 3DResNet, sequence information for 16 frames as input was used. In the 2D-CNN of the 2D CNN-LSTM and ECO, the feature information for the first 16 frame inputs was stored in a buffer, and only the feature information for the $k$th frame was updated in the form of a queue. In the 3D-ResNet, it waited until the first 16 frames were stored in the input buffer to perform inference. Then, in the form of a cue, the first and last frames of the input sequence were replaced to perform inference. Finally, the current surgical phase was determined as the class with the maximum confidence value, and augmentation techniques other than the input image size resizing were not applied in consideration of the real-time scenario [31]. 

\smallskip\noindent \textbf{Results of laparoscopic cholecystectomy.} Table \ref{tab1} shows the results of the surgical phase recognition for each split of cholecystectomy. The cross-validation set consists of 20 training surgical videos and 4 testing surgical videos.  All test performance was measured as the average precision (AP) of the inference results. The test performance of two trained models with different initial weights was averaged for a fair comparison. From the table, it can be seen that the revised annotation (Annotations 2 and 3) shows improved generalization performance compared to the original annotation in 3D-CNN-based surgical phase recognition models. This seems to alleviate the bias of the training information for the part where the classification of the surgical phase is ambiguous (such as the section where the change in the surgical phase occurs) in the process of agreeing with experts. Furthermore, the 3D-CNN-based models show improved generalization performance in all experiments compared to the 2D CNN-LSTM, and it is confirmed that simultaneously utilizing spatiotemporal information in feature processing is effective for the recognition of the surgical phase.

\smallskip\noindent \textbf{Results of robotic subtotal gastrectomy.} We performed cross-validation by dividing 40 cases of gastric cancer surgery into 35 cases of training data and 5 cases of evaluation data.  In order to properly distribute the difficulty of evaluation data, we constructed 3 types of split cross-validation sets by considering the patient's age, operation time, vertebral bleeding amount, and BMI among 40 cases of surgical video. Table \ref{tab2} shows the inference performance for each evaluation split of the surgical phase recognition model trained from the annotations performed by four specialists. At this time, {\it Con} denotes an annotation in which the opinions of the investigator agreed upon. {\it Con-Ann} denotes the difference between the evaluation performance of the model trained with the annotations obtained from each annotator and the performance evaluated by the consensus annotation.

\begin{figure}[t!]
\begin{center}
\includegraphics[width=0.6\linewidth]{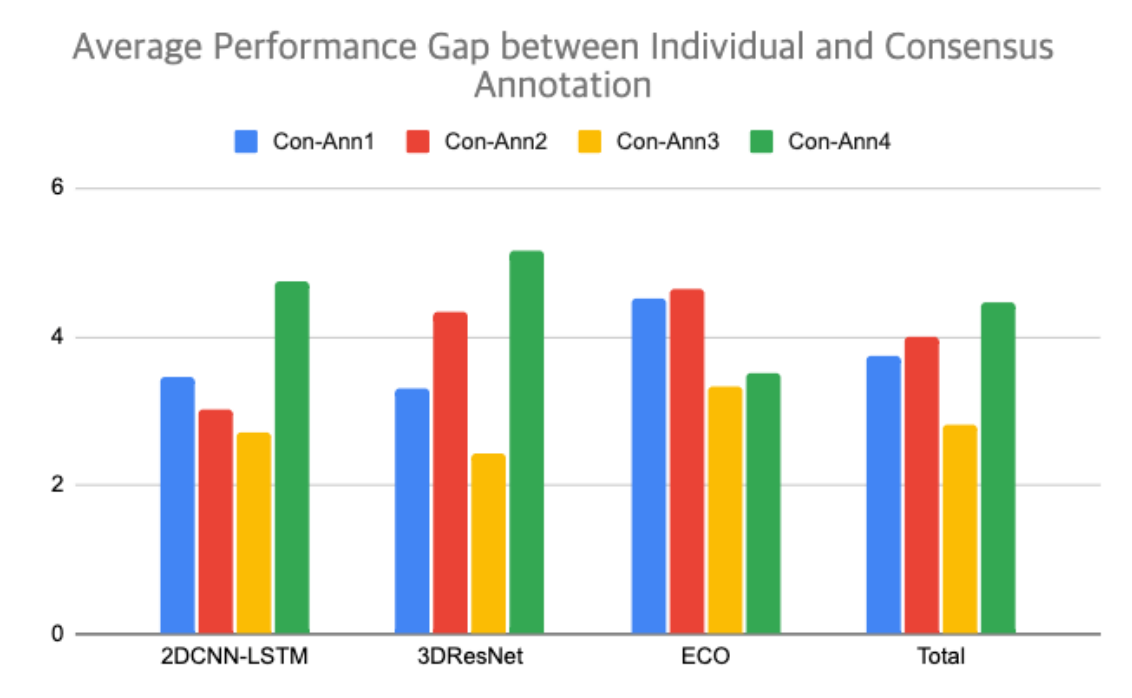}
\caption{\textbf{Average performance gap of all splits between surgical phase recognition model trained with individual and consensus annotation.} Depending on the model, the graph shows a performance gap from a minimum of $2\%$ to a maximum of $5\%$.} \label{fig6}
\end{center}
\end{figure}

\begin{figure}[t!]
\begin{center}
\includegraphics[width=0.9\linewidth]{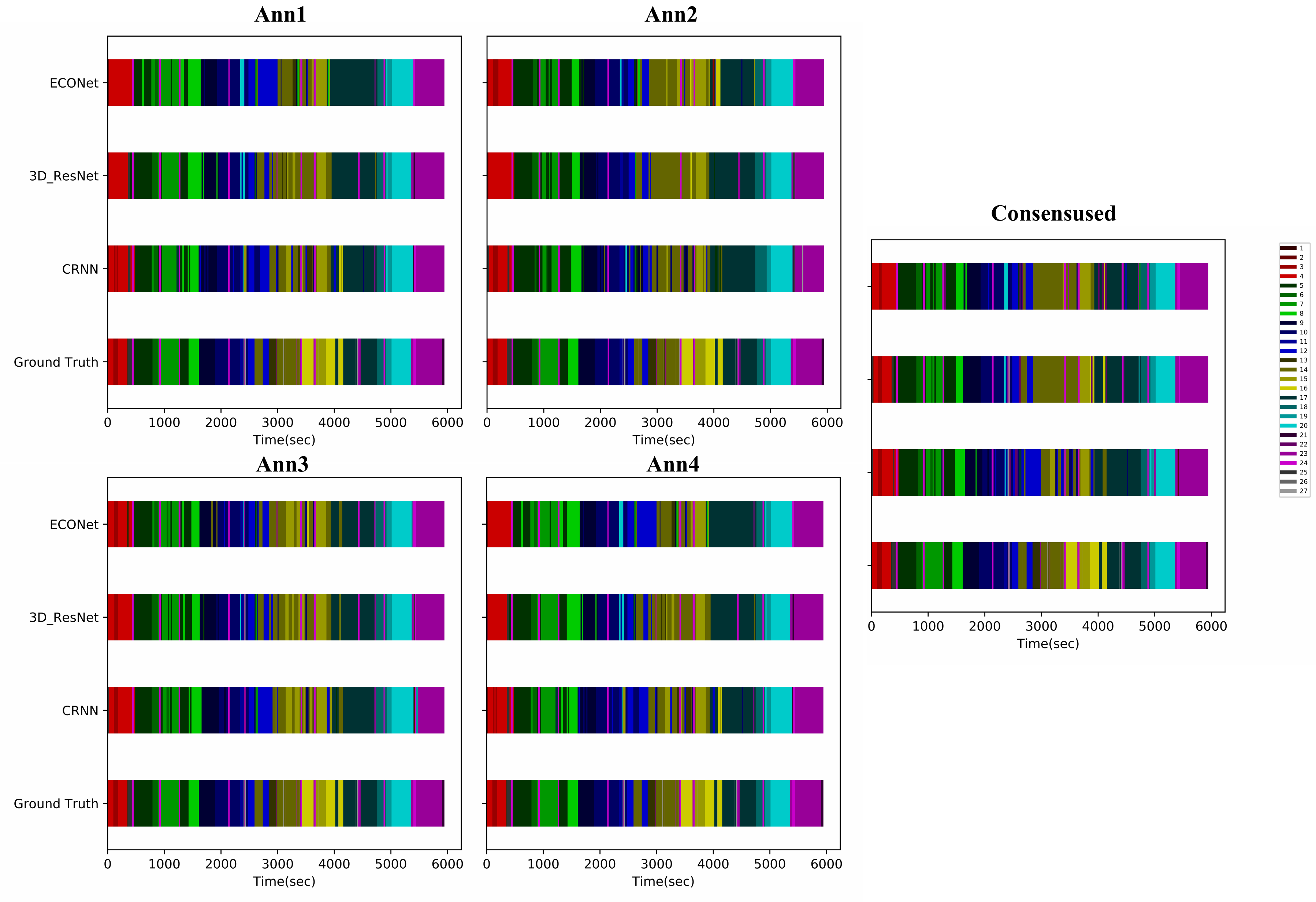}
\caption{\textbf{Visualization of inference output of same surgical video with different annotations.} The visualizations obtained from the model trained by the four different types of annotation and the consensus annotation are shown. bars are colored according to a total of 21 gastrectomy phases and 6 non-gastrectomy actions.} \label{fig7}
\end{center}
\end{figure}

In the case of the surgical phase recognition model trained with consensus annotation, the highest performance was achieved in all models and evaluation splits. In Figure \ref{fig6}, in all splits, the evaluation results varied from at least $2\%$ to $5\%$ depending on the model. Figure \ref{fig7} shows the visualization of the inference results of a model trained with different annotations on the same surgical video. The model trained from the annotations obtained through consensus was able to confirm that the model showed more robust performance in the section with frequent step movements in the middle of the operation, and relatively stable inference results compared to the model trained on each annotation.

\section{Conclusion}

To rethink the generalization performance of the surgical phase recognition, we performed verification on a training model trained with annotations generated by experts. We performed cross-validation on the public Cholec80 dataset [17] and also tried additional validation to subtotal gastrectomy for gastric cancer with multiple expertise annotations. It can be confirmed that the difference in generalization error is probably very large under the same conditions depending on how the annotation process is designed when the difficulty of the annotation is extremely high. Through this study, it was shown that recognition problems with high annotation difficulty, such as surgical video or medical imaging analysis, have to undergo a more rigorous annotation process considering the application of the clinical field. In the future, we plan to conduct research by developing algorithms to secure annotations that can improve generalization performance through model uncertainty or evolutionary algorithms.

\section*{References}


{
\small

[1] Jain, A. K.\ \& Chandrasekaran, B.\ (1982) 39 Dimensionality and sample size considerations in pattern recognition practice. {\it Handbook of Statistics}, {\bf 2}:835-855. 

[2] Raudys, S. J.\ \& Jain, A. K.\ (1991) Small sample size effects in statistical pattern recognition: Recommendations for practitioners. {\it IEEE Transactions on Pattern Analysis and Machine Intelligence}, {\bf13}(3):252-264.

[3] Figueroa, R. L.\ Zeng-Treitler, Q.\ Kandula, S.\ \& Ngo, L. H.\ (2012) Predicting sample size required for classification performance. {\it BMC Med. Inform. Decis. Mak.} {\bf12}(8).

[4] Beleites, C.\ Neugebauer, U.\ Bocklitz, T.\ Krafft, C.\ \& Popp, J.\ (2013) Sample size planning for classification models. {\it Analytica Chimica Acta}, {\bf760}(14): 25-33. 

[5] Deng, J.\ Dong, W.\ Socher, R.\ Li, L. -J.\ Li, K.\ \& Fei-Fei, L.\ (2009) ImageNet: A large-scale hierarchical image database. {\it Proc. of CVPR}.

[6] Lin, T. -Y.\ Maire, M.\ Belongie, S.\ Hays, J.\ Perona, P.\ Ramanan, D.\ Dollar, P.\ \& Zitnick, C. L.\ (2014) Microsoft COCO: Common objects in context. {\it Proc. of ECCV}.

[7] Vondrick, C.\ Patterson, D.\ \& Ramanan, D.\ (2012) Efficiently scaling up crowdsourced video annotation. {\it International Journal of Computer Vision} {\bf101}(1):184-204.

[8] Heilbron F. C.\ \& Niebles, J. C.\ (2014) Collecting and annotating human activities in web videos. {\it Proc. of ICMR}.

[9] Damen, D.\ Doughty, H.\ Farinella, G. M.\ Fidler, S.\ Furnari, A.\ Kazakos, E.\ Moltisanti, D.\ Munro, J.\ Perrett, T.\ Price, W.\ \& Wray, M.\ (2018) Scaling egocentric vision: The EPIC-KITCHENS dataset. {\it Proc. of ECCV}. 

[10] Grunberg, K.\ Jimenez-del-Toro, O.\ Jakab, A.\ Langs, G.\ Fernandez, T. S.\ Winterstein, M.\ Weber, M. -A.\ \& Krenn, M.\ Annotating medical image data. {\it Cloud-Based Benchmarking of Medical Image Analysis} 45-67.  

[11] Pelka, O.\ Nensa, F.\ \& Friedrich, C. M.\ (2018) Annotation of enhanced radiographs for medical image retrieval with deep convolutional neural networks. {\it PLoS ONE} {\bf 13}(11).

[12] Seiferta, S.\ Kelma, M.\ Moellerb, M.\ Mukherjeec, S.\ Cavallarod, A.\ Hubera, M.\ \& Comaniciu, D. \ (2010) Semantic annotation of medical images., {\it Proc. of SPIE Medical Imaging}. 

[13] Lundervold, A. S.\ \& Lundervold, A.\ (2018) An overview of deep learning in medical imaging focusing on MRI. {\it Zeitschrift für Medizinische Physik} {\bf 29}(2):102-127.

[14] Philbrick, K. A.\ Weston, A. D.\ Akkus, Z.\ Kline, T. L.\ Korfiatis, P.\ Sakinis, T.\ Kostandy, P.\ Boonrod, A.\ Zeinoddini, A.\ Takahashi, N.\ \&  Erickson, B. J.\ (2019) RIL-contour: A medical imaging dataset annotation tool for and with deep learning. {\it Journal of Digital Imaging} {\bf 32}(4):571-581.

[15] Kohli, M. D.\ Summers, R. M.\ \& Geis, J. R.\ (2017) Medical image data and datasets in the era of machine learning-whitepaper from the 2016 C-MIMI meeting dataset session. {\it Journal of Digital Imaging} {\bf 30}(4):392-399.

[16] {\O}rting, S.\ Doyle, A.\ van Hilten, A.\ Hirth, M.\ Inel, O.\ Madan, C. R.\ Mavridis, P.\ Spiers, H.\ \& Cheplygina, V.\ (2020) A survey of crowdsourcing in medical image analysis. {\it Human Computation} {\bf 7}(1):1-26.

[17] Twinanda, A. P.\ Shehata, S.\ Mutter, D.\ Marescaux, J.\ de Mathelin, M.\ \& Padoy, N.\ (2017) EndoNet: A deep architecture for recognition tasks on laparoscopic Videos. {\it IEEE Transactions on Medical Imaging} {\bf 36}(1):86-97.

[18] Yu, T.\ Mutter, D.\ Marescaux, J.\ \& Padoy, N.\ (2018) Learning from a tiny dataset of manual annotations: a teacher/student approach for surgical phase recognition. {\it Proc. of IPCAI}.

[19] Yengera, G.\ Mutter, D.\ Marescaux, J.\ \& Padoy, N.\ (2018) Less is more: Surgical phase recognition with less annotations through self-supervised pre-training of CNN-LSTM networks. {\it arXiv:1805.08569}.

[20] Loukas, C.\ (2019) Surgical phase recognition of short video shots based on temporal modeling of deep features. {\it Proc. of BIOSTEC}.

[21] Kitaguchi, D.\ Takeshita, N.\ Matsuzaki, H.\ Takano, H.\ Owada, Y.\ Enomoto, T.\ Oda, T.\ Miura, H.\ Yamanashi, T.\ Watanabe, M.\ Sato, D.\ Sugomori, Y.\ Hara, S.\ \& Ito, M.\ (2020) Real-time automatic surgical phase recognition in laparoscopic sigmoidectomy using the convolutional neural network-based deep learning approach.  {\it Surgical Endoscopy} {\bf 34}(11):4924-4931.

[22] Meeuwsen, F. C.\ van Luyn, F.\ Blikkendaal, M. D.\ Jansen, F. W.\ \& van den Dobbelsteen, J. J.\ (2019) Surgical phase modeling in minimal invasive surgery. {\it Surgical Endoscopy} {\bf 33}(5):1426-1432. 

[23] Zia, A.\ Hung, A.\ Essa, I.\ \& Jarc, A.\ (2018) Surgical activity recognition in robot-assisted radical prostatectomy using deep learning. {\it Proc. of MICCAI}.

[24] Kong, Y.\ \& Fu, Y.\ (2018) Human action recognition and prediction. {\it CoRR abs/1806.11230}.

[25] Bhoi, A.\ (2019) Spatio-temporal action recognition: A survey. {\it CoRR abs/1901.09403}.

[26] Simonyan, K.\ \& Zisserman, A.\ (2014) Two-stream convolutional networks for action recognition in videos. {\it In Proc. of NIPS}.

[27] Donahue, J.\ Hendricks, L. A.\ Rohrbach, M.\ Venugopalan S.\ Guadarrama, S.\ Saenko, K.\ \& Darrell, T.\ (2015) Long-term recurrent convolutional networks for visual recognition and description. {\it Proc. of CVPR}.

[28] Tran, D.\ Wang, H.\ Torresani, L.\ Ray, J.\ LeCun, Y.\ \& Paluri, M.\ (2018) A closer look at spatiotemporal convolutions for action recognition, {\it Proc. of CVPR}.

[29] Hara, K.\ Kataoka, H.\ \& Satoh, Y.\ (2018) Can spatiotemporal 3D CNNs retrace the history of 2D CNNs and ImageNet? {\it Proc. of CVPR}.

[30] Feichtenhofer, C.\ Fan, H.\ Malik, J.\ \& He, K.\ (2019) SlowFast networks for video recognition. {\it Proc. of ICCV}.

[31] Zolfaghari, M.\ Singh, K.\ \& Brox, T.\ (2018) ECO: Efficient Convolutional Network for Online Video Understanding. {\it Proc. of ECCV}.

[32] Lee, J. H.\ Tanaka, E.\ Woo, Y.\  Ali, G.\ Son, T.\ Kim, H., -I.\ \& Hyung, W., J.\ (2017) Advanced real-time multi-display educational system (ARMES): An innovative real-time audiovisual mentoring tool for complex robotic surgery. {\it Journal of Surgical Oncology} {\bf 116}(7):894-897.

[33] Paszke, A.\ Gross, S.\ Massa, F.\ Lerer, A.\ Bradbury, J.\ Chanan, G.\ Killeen, T.\ Lin, Z.\ Gimelshein, N.\ Antiga, L.\ Desmaison, A.\ Kopf, A.\ Yang, E.\ DeVito, Z.\ Raison, M.\ Tejani, A.\ Chilamkurthy, S.\ Steiner, B.\ Fang, L.\  Bai, J.\ \& Chintala. S.\ (2019) PyTorch: An imperative style, high-performance deep learning library. {\it Proc. of NeurIPS}.

[34] He, K.\ Zhang, X.\ Ren, S.\ \& Sun, J.\ (2016) Deep residual learning for image recognition. {\it Proc. of CVPR}.

[35] Kay, W.\ Carreira, J.\ Simonyan, K.\ Zhang, B.\ Hillier, C.\ Vijayanarasimhan, S.\ Viola, F.\ Green, T.\ Back, T.\ Natsev, P.\ Suleyman, M.\ \& Zisserman, A.\ (2017) The Kinetics human action video dataset. {\it CoRR abs/1705.06950}.

[36] Szegedy, C.\ Ioffe, S.\ Vanhoucke, V.\ \&  Alemi A.\ (2017) Inception-v4, Inception-ResNet and the impact of residual connections on learning. {\it Proc. of AAAI}.

\section*{Supplementary Materials}

\section*{A. Network Architecture}
Figure \ref{fig1_supp} shows schematics of model architectures of 2D-CNN-LSTM [27], 3D-ResNet [29], and ECO [31] models. Test scenarios were constructed according to the structural characteristics of the trained model, and no pre- or post-processing of the input data and inference output was performed to accurately analyze the effect of the label on the model training. However, in the training and inference of each CNN model, basic data augmentation was performed as part of the regularization at the input stage of the architecture.  

\noindent\textbf{Annotation revision for laparoscopic cholecystectomy.} We revised the annotation of the phase of the laparoscopic cholecystectomy surgery through a consensus process of three specialists. The consensus among specialists for the seven steps of cholecystectomy was based on their consensus, and a separate consensus process was performed on the points where changes in the surgical phase occurred or where exceptions occurred. Finally, two types of annotation (Annotation 2 and 3) revised by two specialists were created from Annotation 1. Table \ref{tab1_supp} shows the details of each redefined surgical phase. \\

\noindent\textbf{Annotation pipeline for robotic subtotal gastrectomy.} For accurate labeling for more complex surgery, we performed frame-level annotation for 40 cases that were cross-validated by three specialists. For cross-validation, we received annotated information according to each opinion from four specialists, and this information was compared with the areas where the opinions matched by AND operation. For labels containing blank generated by the AND operation, another specialist annotates the label as an investigator, resulting in the final consensus label. Robotic surgery for 40 cases of gastric cancer proceeded subtotal gastrectomy and nineteen cases were recorded on the da Vinci Si system, while twenty-one videos were recorded on the da Vinci Xi. Table \ref{tab3_supp} shows the details of each surgical phase of subtotal gastrectomy.

\begin{figure}[h!]
\includegraphics[width=\textwidth]{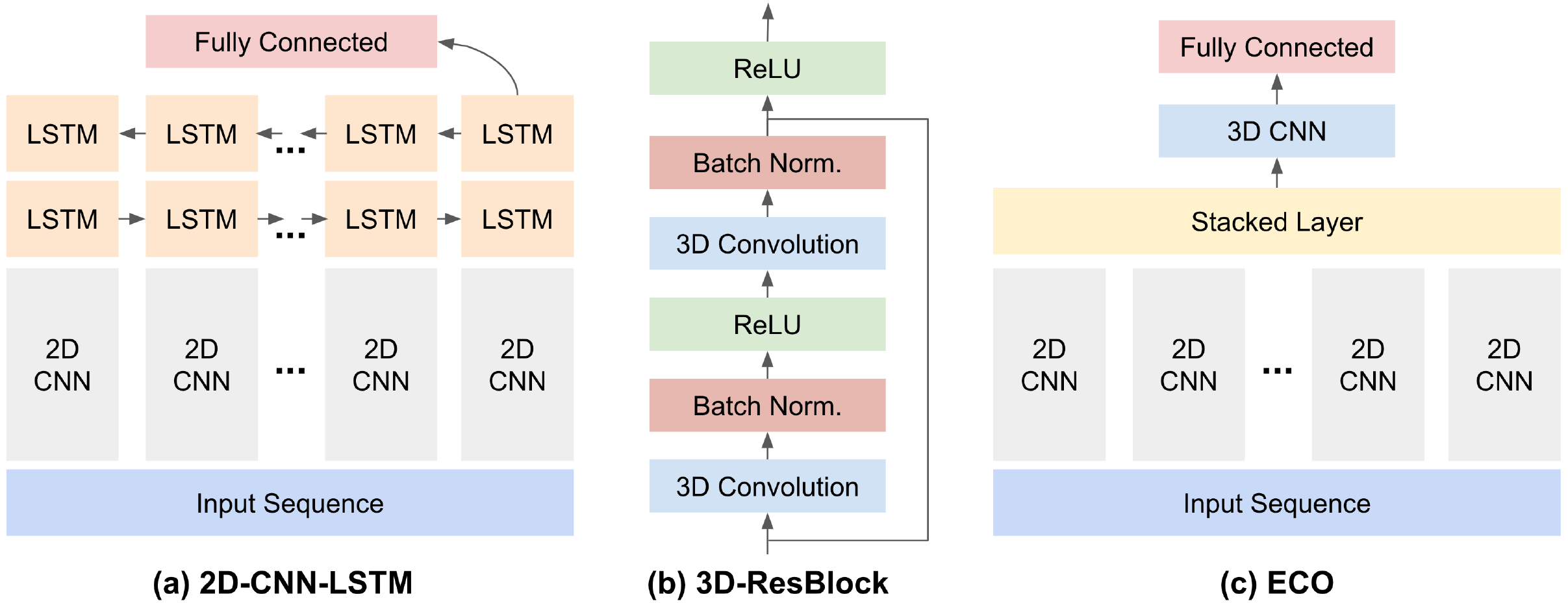}
\caption{\textbf{Schematic illustration of each network architecture utilized for surgical phase recognition.} 2D-CNN-LSTM encodes information in the spatial domain with 2D-CNN only, and then considers the information together in the time domain, and the other two models simultaneously consider spatiotemporal information through the 3D convolution layer.} \label{fig1_supp}
\end{figure}

\begin{table*}[h!]
  \caption{Surgical phase of cholecystectomy redefined through expert consensus.}
  \label{tab1_supp}
  \begin{center}
  \scalebox{0.8}{
  \begin{tabular}{c|c}
\hline
  \textbf{Phase ID} & \textbf{Definition} \\ \hline \hline
preparation (0) & \makecell{Time period from the start of the surgery until the first laparoscopic tool appears.} \\ \hline 
calot triangle dissection (1) & \makecell{Time period from when the first laparoscopic tool appears \\ until the first clip applier appears.} \\ \hline
clipping and cutting (2) & \makecell{Time period from when the first clip applier appears until the cystic.a \& duct is finished \\ (The end point is until the next tool touches the organization).} \\ \hline

gallbladder dissection (3) & Time period from when the tool touches tissue until the specimen bag appears. \\ \hline 
gallbladder packaging (4) & \makecell{Time period from when the specimen bag appears until it is closed \\ (or until a suction tip or other aid tool appears).} \\ \hline
cleaning and coagulation (5) &  \makecell{Time period from when the specimen bag is closed until the tool touches the bag strap \\ (or until the suction tip is withdrawn or the trocar appears on the screen).} \\ \hline
gallbladder retraction (6) &  \makecell{Time period from the first touch of the bag strap until the specimen bag is withdrawn.} \\ 
\hline
\end{tabular}
}
\end{center}
\end{table*}

\begin{table*}[b!]
  \caption{\textbf{Defining the surgical stage of subtotal gastrectomy for gastric cancer based on ARMES.} IDs 1 to 21 include surgical steps related to gastrectomy, and 22 to 27 refer to non-gastrectomy actions not included in the subtotal gastrectomy process.}
  \label{tab2_supp}
  \begin{center}
  \scalebox{0.9}{
  \begin{tabular}{|c|c|c|c|}
\hline
  \textbf{ID} & \textbf{ Surgical phase} & \textbf{ID} & \textbf{ Surgical phase} \\ \hline \hline
1 & Trocar insertion & 15 & \makecell{Dissection of LN station 7 \\ and ligation of left gastric artery} \\ \hline 
2 & Docking & 16 & Dissection of LN station 11p \\ \hline 
3 & \makecell{Division of less omentum \\ up to the right side of the esophagus} & 17 & \makecell{Clearance of soft tissue \\ along the lesser curvature} \\ \hline 
4 & Liver retraction & 18 & Gastric transection \\ \hline 
5 & Partial (or total) omentectomy & 19 & Harvesting resected  specimen into Endo bag \\ \hline 
6 & Ligation of left gastroepiploic vessels & 20 & Anastomosis \\ \hline 
7 & \makecell{Clearance of soft tissues \\ along the greater curvature} & 21 & Retrieval of specimen \\ \hline 
8 & Ligation of Right gastroepiploic vein & 22 & Adhesiolysis \\ \hline 
9 & Ligation of Right Gastroepiploic Artery & 23 & Housekeeping \\ \hline 
10 & Creation of window for duodenal transection & 24 & Clean camera \\ \hline 
11 & Duodenal transection & 25 & Junk \\ \hline 
12 & Ligation of right gastric artery & 26 & Other procedure \\ \hline 
13 & Dissection of LN stations 12a & 27 & Unexpected surgical events \\ \hline 
14 & Dissection of LN station 8 and 9 & - & - \\ \hline 
\end{tabular}
}
\end{center}
\end{table*}

\section{B. Additional Results}
Table \ref{tab3_supp} shows per phase recognition performance. As seen in Table 3, 3D-CNN based model outperforms every phase in laparoscopic cholecystectomy by 2D-CNN-LSTM model, and 3D-ResNet achieved the highest performance in 12 phases for subtotal gastrectomy.

\noindent\textbf{Additional statistics for laparoscopic cholecystectomy.} Figure \ref{fig2_supp}, \ref{fig3_supp}, and \ref{fig4_supp} show the confusion matrix for each surgical video in split 1 for each training model trained with different annotations. These figures shows the relationship between classes that are difficult to accurately classify. Figures \ref{fig5_supp}, \ref{fig6_supp}, and \ref{fig7_supp} visualize the inferred surgical phase on the time axis versus the ground truth label for the surgical phase. In all graphs, the ground truth label is orange and the inferred output is blue. Models trained at different annotations can be observed on the time base to indicate which inference output. In the case of the surgical phase recognition model trained with consensus annotation, the highest performance was achieved in all models and evaluation splits. 

\noindent\textbf{Additional statistics for robotic subtotal gastrectomy.} Figure \ref{fig8_supp}, and \ref{fig9_supp} show the confusion matrix for two example cases in split 1 for each training model trained with consensus annotations, respectively. Figure \ref{fig10_supp}, and \ref{fig11_supp} show the mean AP for each split and surgical phase and the deviation value of each mean AP in laparoscopic cholecystectomy and robotic gastrectomy, respectively. The graph in Figure 5 and 6 show the sensitivity of generalization to each model.

\begin{table}[b!]
  \caption{Generalization AP for each surgical phase for laparoscopic cholecystectomy and robotic gastrectomy.}
  \label{tab3_supp}
  \resizebox{\linewidth}{!}{
  \begin{tabular}{c|c|c|c|c|c|c|c|c|c|c|c|c|c|c}
\hline
\multicolumn{15}{c}{Laparoscopic cholecystectomy} \\ \hline \hline
Phase ID &  \multicolumn{2}{c|}{0}  &  \multicolumn{2}{c|}{1}  &  \multicolumn{2}{c|}{2}  & \multicolumn{2}{c|}{3} & \multicolumn{2}{c|}{4} &  \multicolumn{2}{c|}{5} & \multicolumn{2}{c}{6} \\ \hline 
2D-CNN-LSTM & \multicolumn{2}{c|}{71.8}  &  \multicolumn{2}{c|}{74.1}  &  \multicolumn{2}{c|}{36.5}  & \multicolumn{2}{c|}{44.7} & \multicolumn{2}{c|}{56.8} & \multicolumn{2}{c|}{57.1} & \multicolumn{2}{c}{37.7} \\ \hline 
3D-ResNet & \multicolumn{2}{c|}{\textbf{78.7}}  &  \multicolumn{2}{c|}{\textbf{76.7}}  &  \multicolumn{2}{c|}{30.6}  & \multicolumn{2}{c|}{37.6} & \multicolumn{2}{c|}{60.1} & \multicolumn{2}{c|}{53.7} & \multicolumn{2}{c}{23.4} \\ \hline 
ECO & \multicolumn{2}{c|}{75.8}  &  \multicolumn{2}{c|}{72.0}  &  \multicolumn{2}{c|}{\textbf{42.3}}  & \multicolumn{2}{c|}{\textbf{57.8}} & \multicolumn{2}{c|}{\textbf{69.2}} & \multicolumn{2}{c|}{\textbf{59.4}} & \multicolumn{2}{c}{\textbf{44.9}} \\ \hline 
\multicolumn{15}{c}{Robotic gastrectomy for gastric cancer} \\ \hline \hline
Phase ID & 1  & 2 & 3 & 4 & 5 & 6 & 7 & 8 & 9 & 10 & 11 & 12 & 13 & 14 \\ \hline 
2D-CNN-LSTM & 64.0 & 58.5 & 31.2 & \textbf{81.2} & 74.1 & 50.2 & 37.9 & \textbf{62.4} & 60.9 & 76.3 & 47.2 & \textbf{39.3} & 0.3 & 23.3 \\ \hline
3D-ResNet & 60.6 & \textbf{69.7} & \textbf{40.5} & 72.8 & \textbf{76.1} & \textbf{55.9} & \textbf{57.8} & 53.7 & 62.5 & 76.4 & 52.1 & 22.7 & 0.4 & \textbf{62.1} \\ \hline
ECO & \textbf{70.3} & 66.8 & 32.2 & 76.6 & 72.7 & 32.9 & 57.4 & 54.5 & \textbf{72.7} & \textbf{82.8} & \textbf{54.0} & 26.4 & \textbf{3.3} & 55.8 \\ \hline
Phase ID & 15  & 16 & 17 & 18 & 19 & 20 & 21 & 22 & 23 & 24 & 25 & 26 & 27 & - \\ \hline
2D-CNN-LSTM & \textbf{49.7} & \textbf{38.5} & 74.1 & 72.5 & 53.8 & \textbf{75.6} & 26.9 & 13.2 & 76.1 & \textbf{59.6} & 33.1 & 0.3 & 0.0 & - \\ \hline
3D-ResNet & 31.9 & 31.2 & 67.4 & 90.1 & \textbf{70.7} & 51.4 & \textbf{53.4} & \textbf{30.8} & 49.0 & 27.8 & \textbf{71.3} & \textbf{5.7} & \textbf{3.8} & - \\ \hline
ECO & 32.0 & 23.2 & \textbf{76.2} & \textbf{91.0} & 54.9 & 62.0 & 10.6 & 25.4 & \textbf{88.6} & 30.0 & 64.0 & 4.0 & 0.0 & - \\ \hline
\end{tabular}
}
\end{table}

\begin{figure}[b!]
\includegraphics[width=\textwidth]{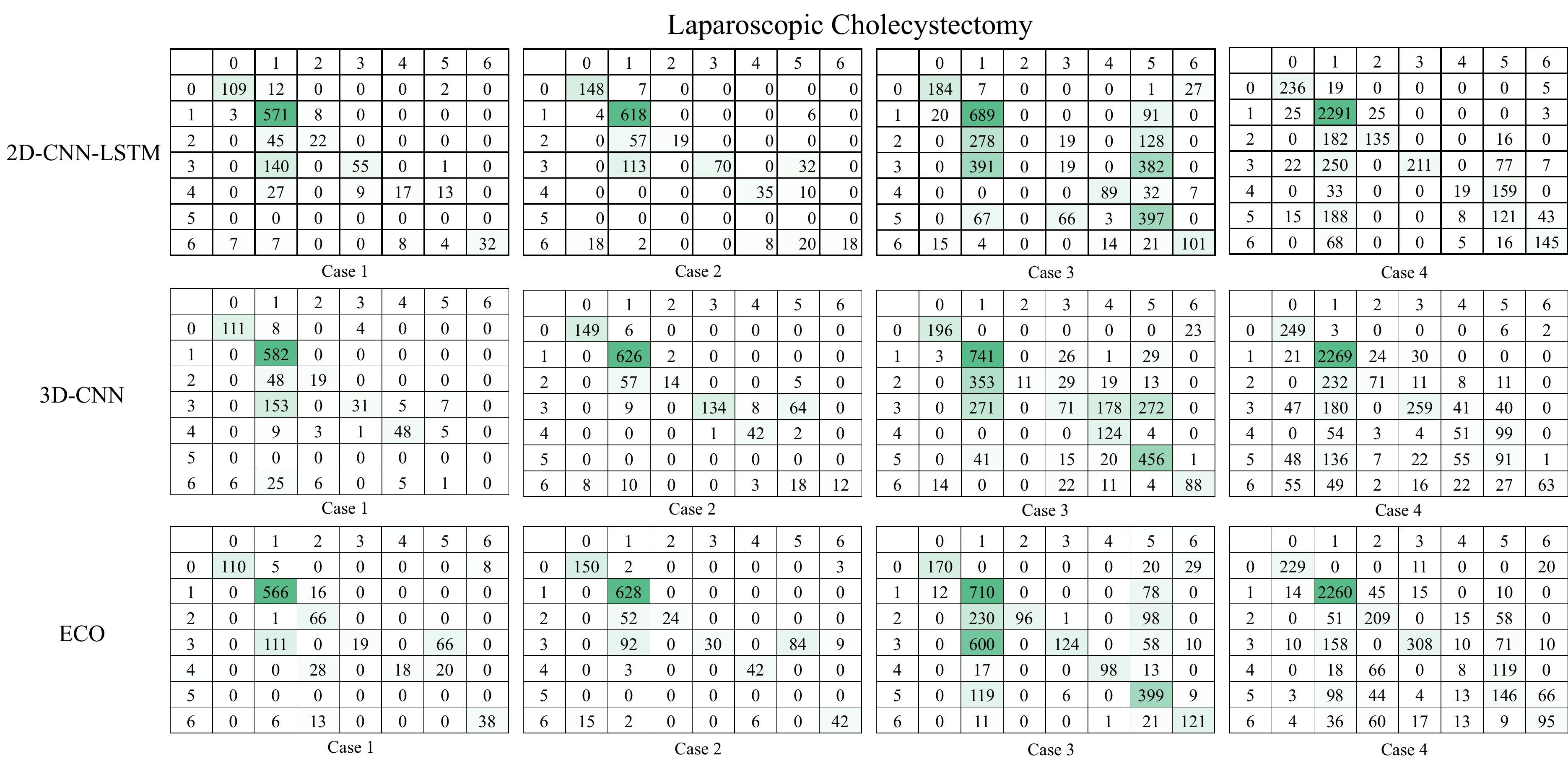}
\caption{\textbf{Confusion matrix for inference output from 2D-CNN-LSTM trained with difference annotations.} The unit of each cell means one input clip in each surgical phase.} \label{fig2_supp}
\end{figure}

\begin{figure}[t!]
\includegraphics[width=\textwidth]{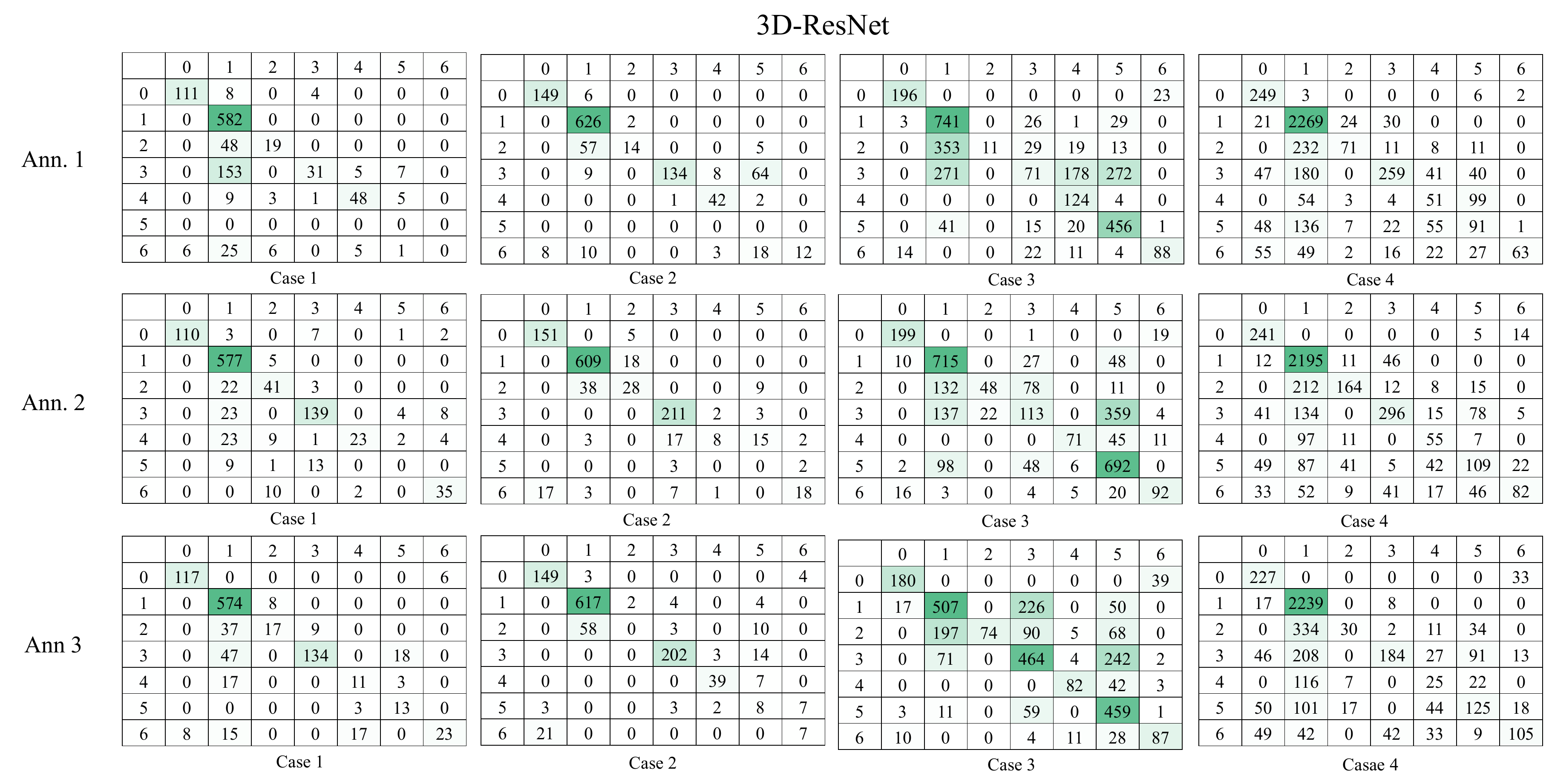}
\caption{\textbf{Confusion matrix for inference output from 3D-CNN trained with difference annotations.} The unit of each cell means one input clip in each surgical phase.} \label{fig3_supp}
\end{figure}

\begin{figure}[t!]
\includegraphics[width=\textwidth]{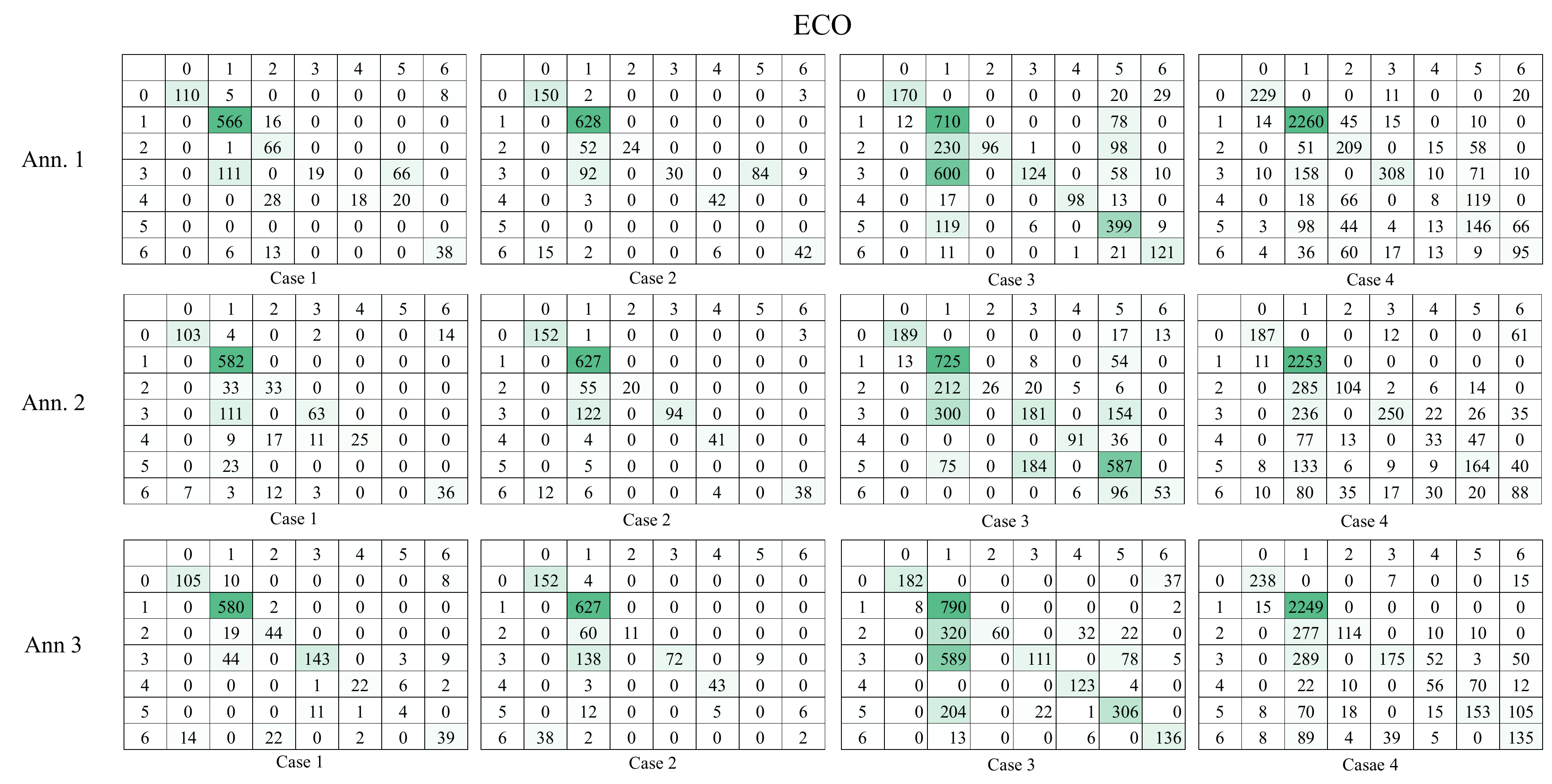}
\caption{\textbf{Confusion matrix for inference output from ECO trained with difference annotations.} The unit of each cell means one input clip in each surgical phase.} \label{fig4_supp}
\end{figure}

\begin{figure}[t!]
\includegraphics[width=\textwidth]{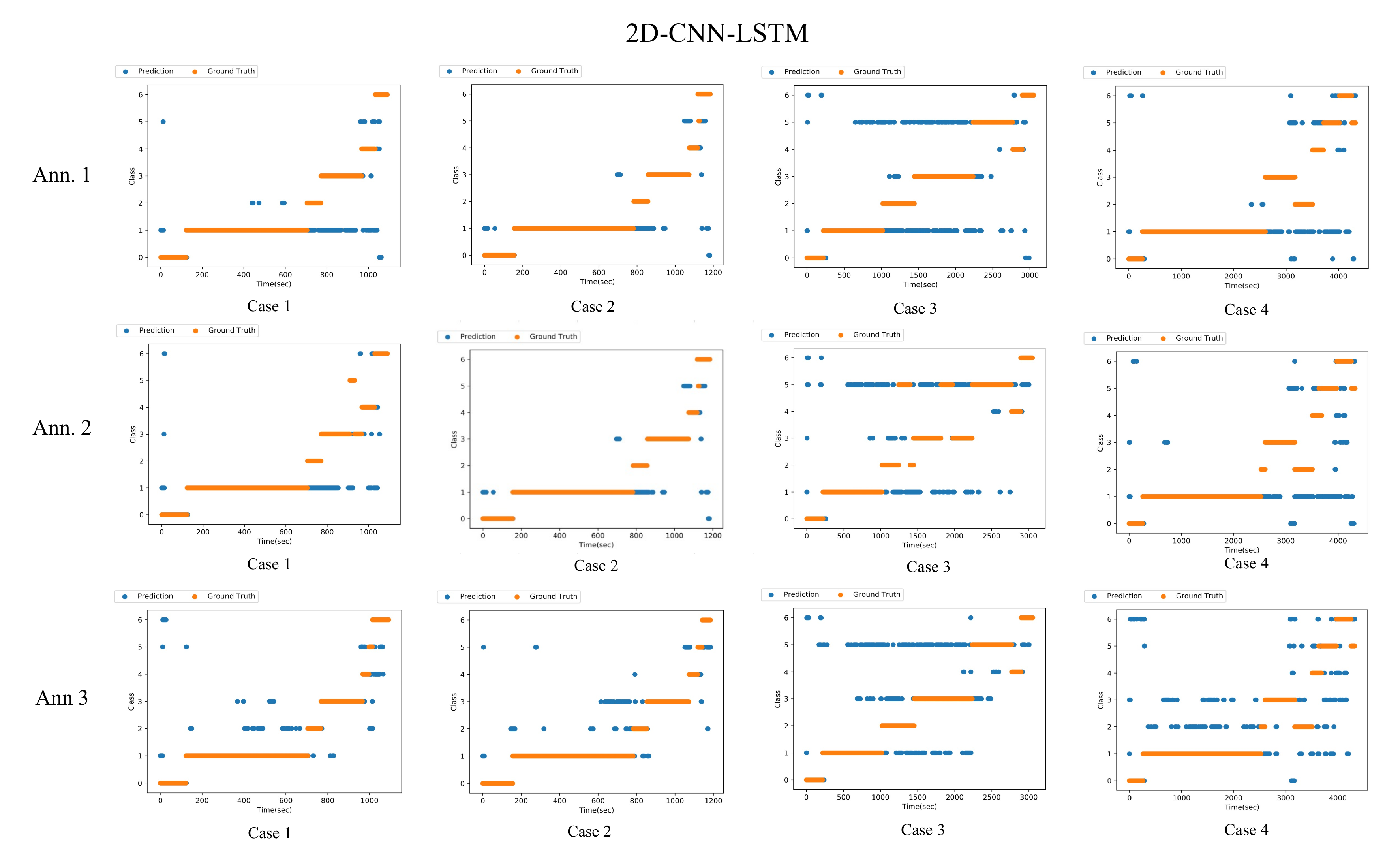}
\caption{\textbf{Visualization on the time axis for inference output from 2D-CNN-LSTM.} In all graphs, orange is the ground truth label and blue is the inference output.} \label{fig5_supp}
\end{figure}

\begin{figure}[t!]
\includegraphics[width=\textwidth]{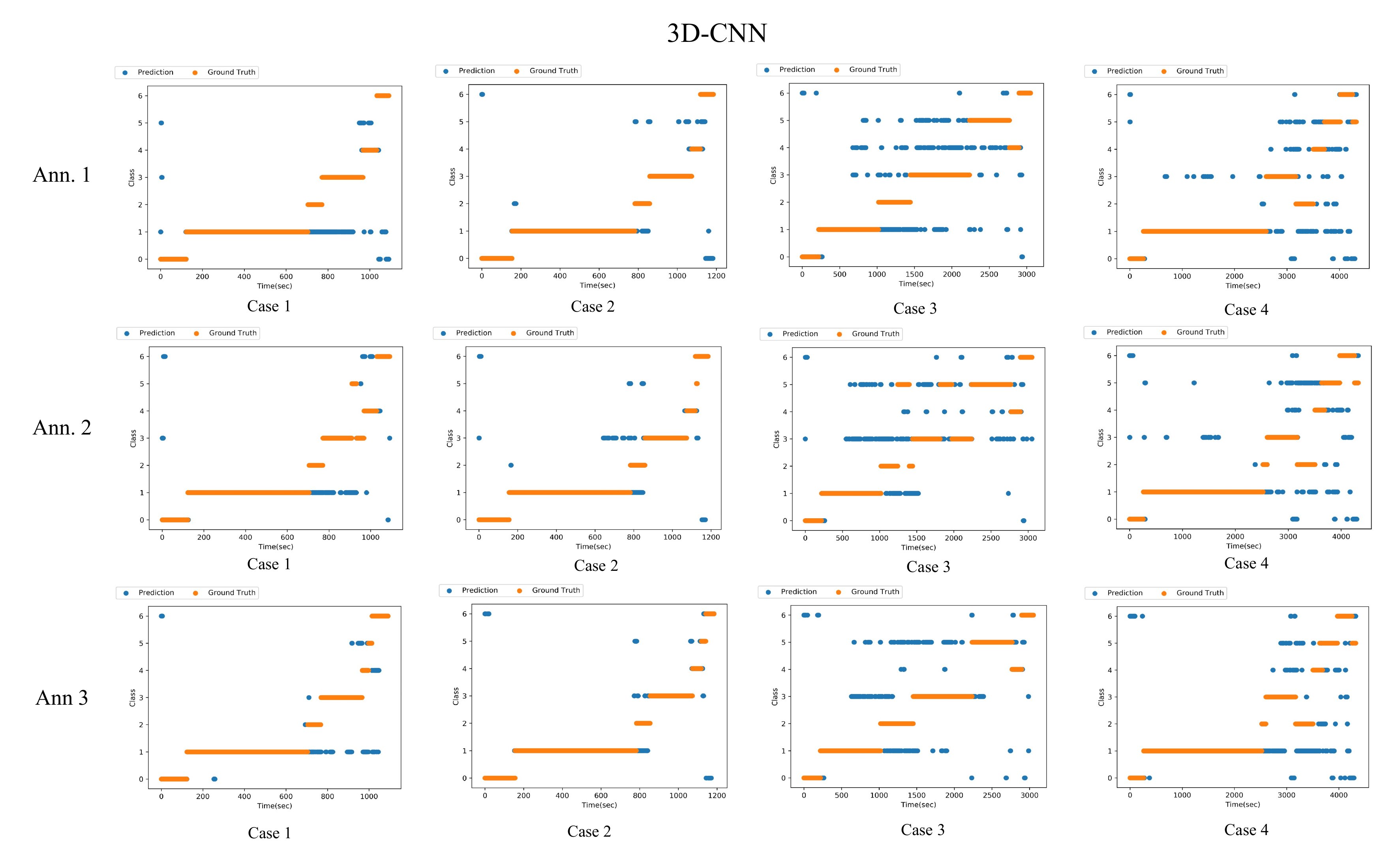}
\caption{\textbf{Visualization on the time axis for inference output from 3D-CNN.} In all graphs, orange is the ground truth label and blue is the inference output.} \label{fig6_supp}
\end{figure}

\begin{figure}[t!]
\includegraphics[width=\textwidth]{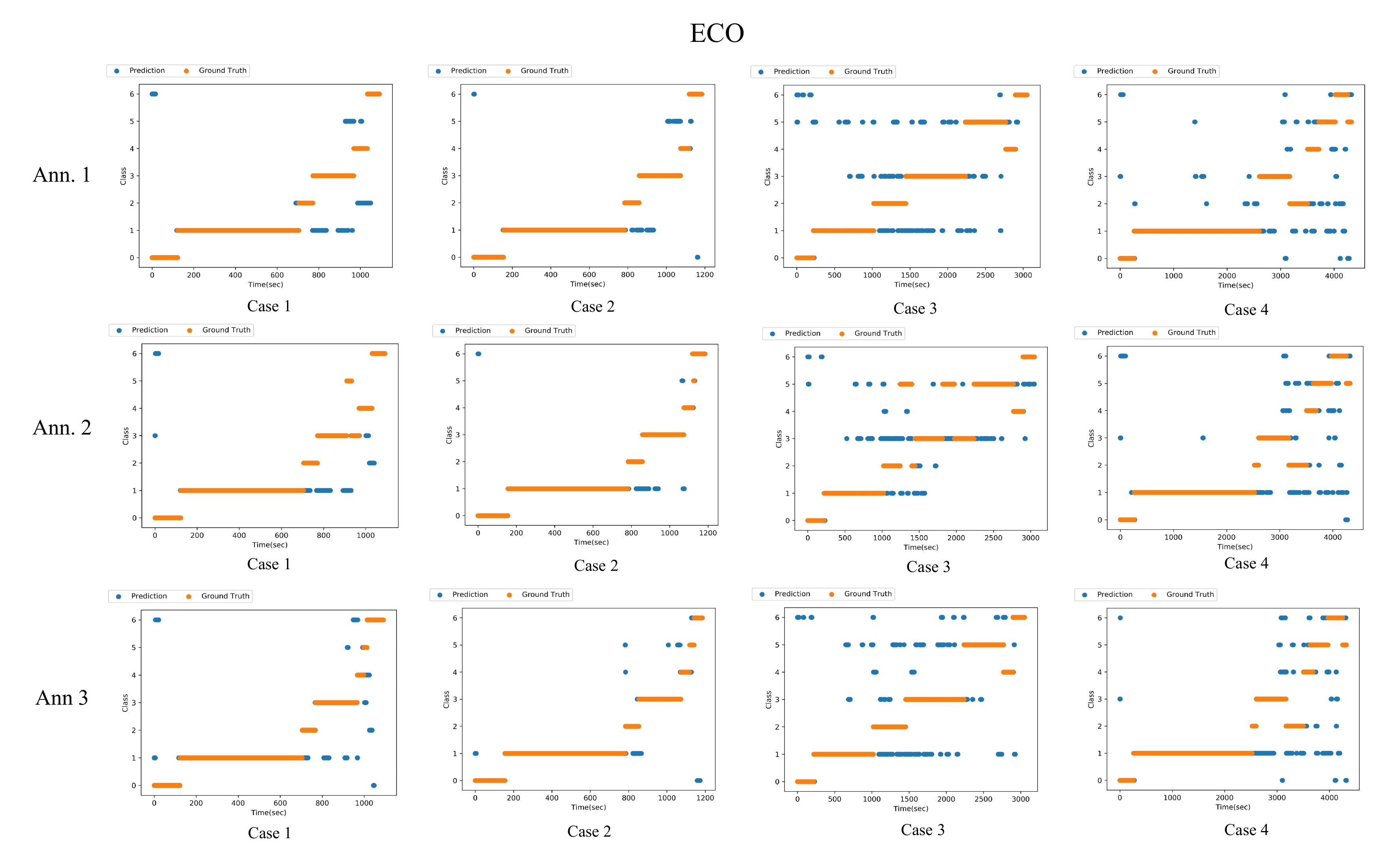}
\caption{\textbf{Visualization on the time axis for inference output from ECO.} In all graphs, orange is the ground truth label and blue is the inference output.} \label{fig7_supp}
\end{figure}

\begin{figure}[t!]
\includegraphics[width=\textwidth]{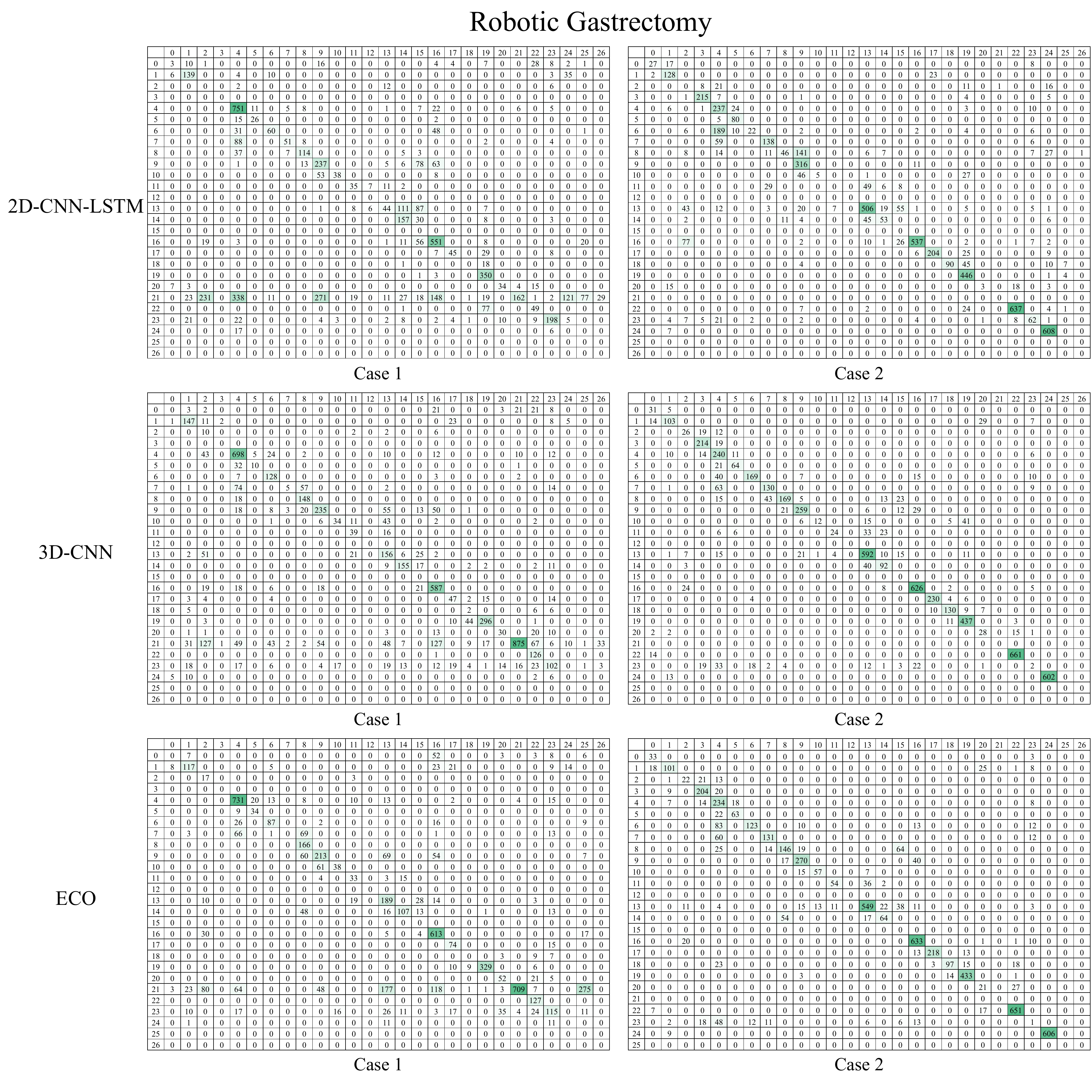}
\caption{\textbf{Confusion matrix for inference output from difference models in robotic subtotal gastrectomy.} The unit of each cell means one input clip in each surgical phase.} \label{fig8_supp}
\end{figure}

\begin{figure}[t!]
\includegraphics[width=\textwidth]{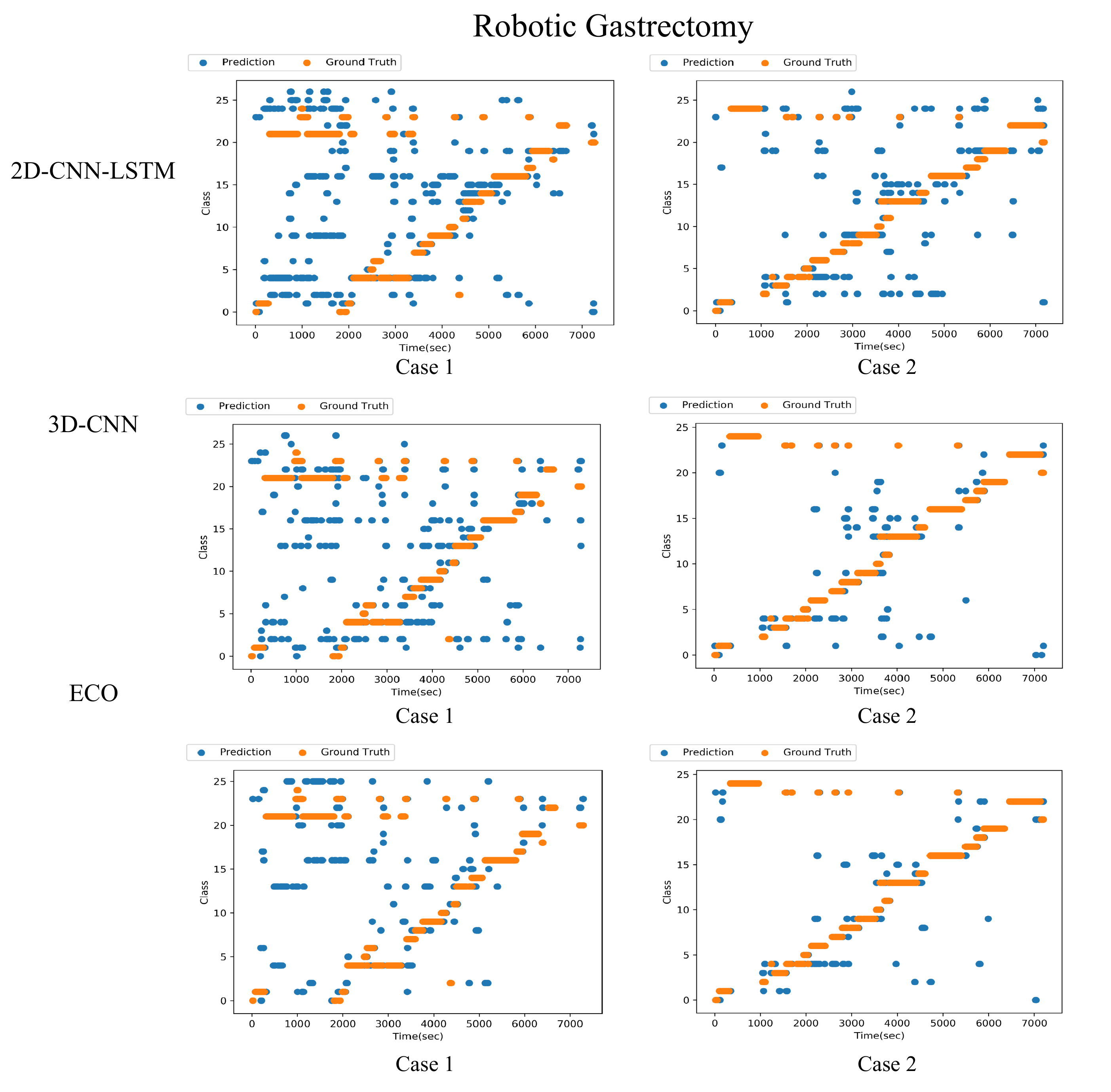}
\caption{\textbf{Visualization on the time axis for inference output for robotic subtotal gastrectomy.} In all graphs, orange is the ground truth label and blue is the inference output.} \label{fig9_supp}
\end{figure}

\begin{figure}[t!]
\includegraphics[width=\textwidth]{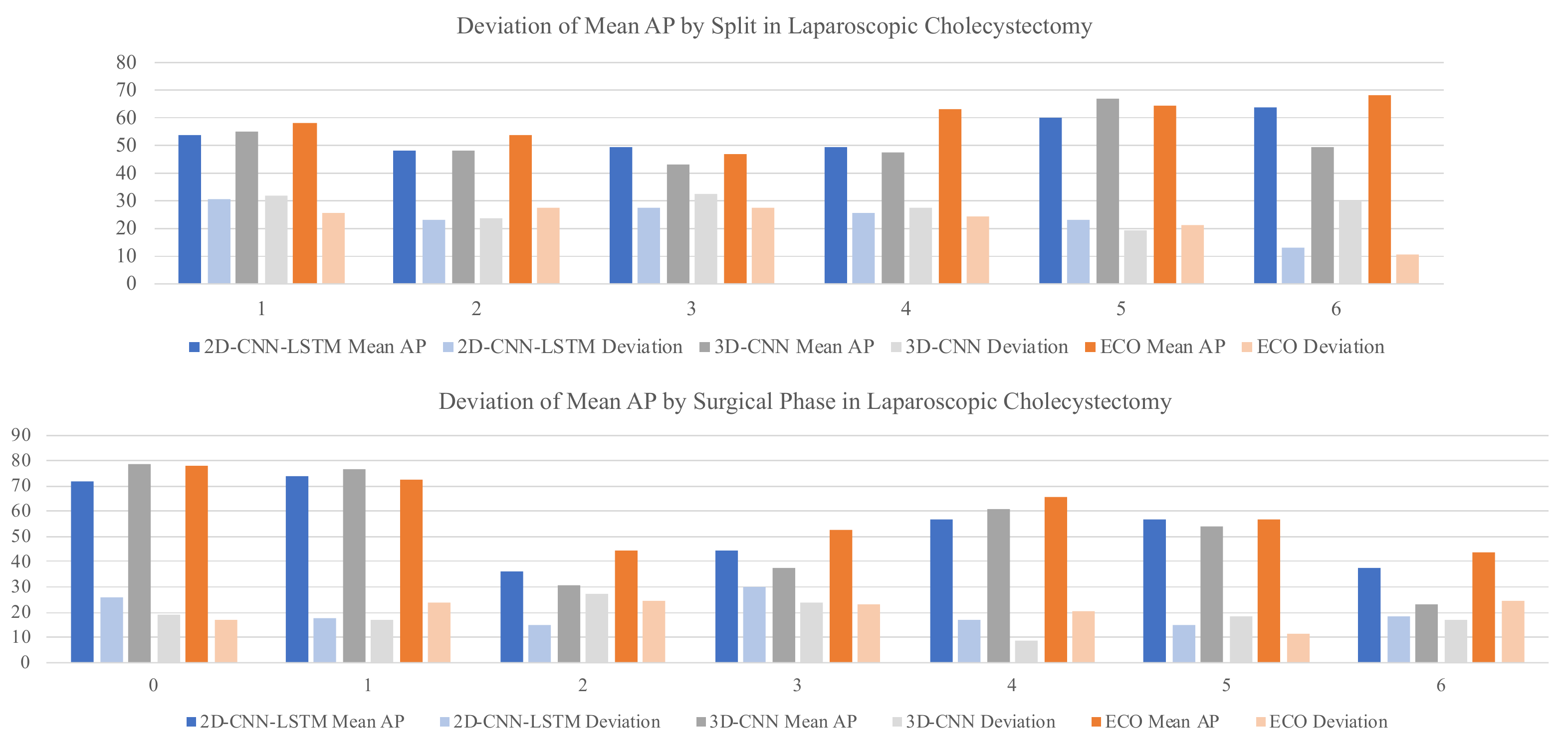}
\caption{\textbf{Deviation of mean AP by model in each split and surgical phase of laparoscopic cholecystectomy.} Dark bars indicate total AP and light bars indicate deviation from Total AP.} \label{fig10_supp}
\end{figure}

\begin{figure}[t!]
\includegraphics[width=\textwidth]{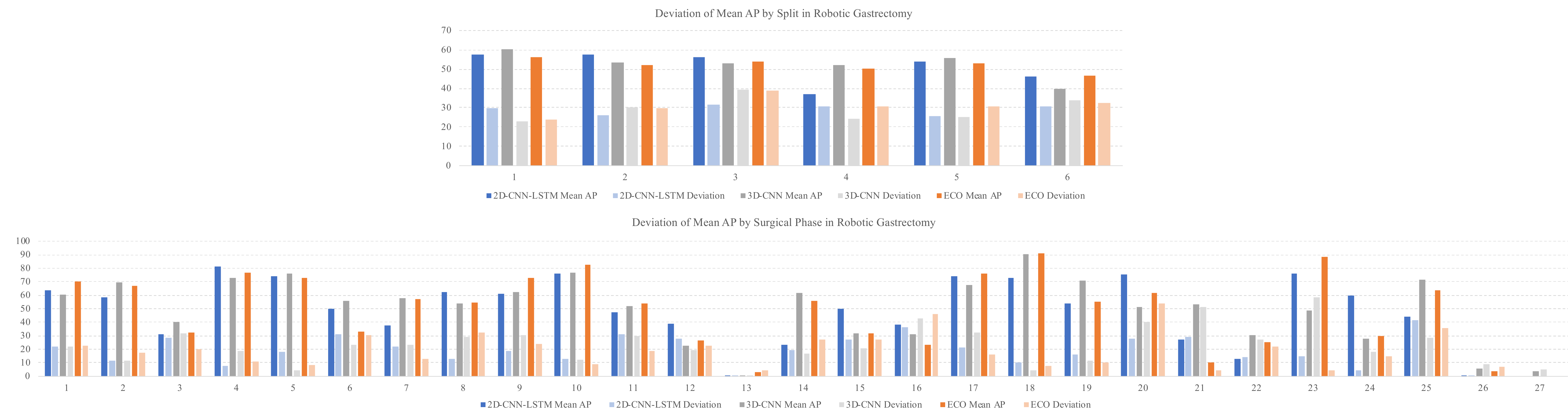}
\caption{\textbf{Deviation of mean AP by model in each split and surgical phase of robotic subtotal gastrectomy.} Dark bars indicate Total AP and light bars indicate deviation from total AP.} \label{fig11_supp}
\end{figure}

\end{document}